%% file: hwe.tex
\documentclass[11pt,a4paper]{article}
\usepackage[hyperref]{conll-2019}
\usepackage{times}
\usepackage{latexsym}
\usepackage{multirow}
\usepackage{graphicx}
\usepackage{amsmath}
\usepackage{subcaption}
\usepackage{amsfonts}
\usepackage{soul}
\usepackage{url}

\aclfinalcopy 

\title{A Probabilistic Framework for Learning Domain Specific\\   Hierarchical Word Embeddings}

\author{Lahari Poddar \\ Amazon \\ poddarl@amazon.com
\And Gy\"orgy Szarvas \\ Amazon \\ szarvasg@amazon.com
\And Lea  Frermann\thanks{\small{*Work done while the author was at Amazon}} \\ University of Melbourne \\ lea.frermann@unimelb.edu.au}

\date{}

\begin{document}
\maketitle
\begin{abstract}
\input{abstract}
\end{abstract}

\input{introduction_v3}

\input{relatedWork_v3}
\input{methods_v6}

\input{experiments_v6}

\input{conclusion_v2}

\newpage
\bibliographystyle{acl_natbib}
\bibliography{references}
\end{document}

%% file: abstract.tex
The meaning of a word often varies depending on its usage in different domains.
The standard word embedding models struggle to represent this variation, as they learn a single global representation for a word.
We propose a method to learn domain-specific
word embeddings, from text organized into hierarchical domains, such as reviews in an e-commerce website, where products follow a taxonomy. Our structured probabilistic model
allows vector representations for the same word to drift away from each other
for distant domains in the taxonomy, to accommodate its domain-specific meanings. 
By learning sets of domain-specific word representations {\it 
jointly}, our model can leverage domain relationships, and it scales well 
with the number of domains.
Using large real-world review datasets, we demonstrate the 
effectiveness of our model compared to state-of-the-art approaches, in learning 
domain-specific word embeddings that are both 
intuitive to humans and benefit downstream NLP tasks.

%% file: introduction_v3.tex
\section{Introduction}
\label{introduction}
Word embedding models learn a lower dimensional vector representation of a word, 
while encoding its semantic relationship to other words in a 
corpus~\cite{mikolov2013efficient}.
Using pre-trained embeddings to represent the semantics of input words has 
become a standard practice in NLP models. Apart from their usage in downstream applications, word embeddings are also powerful 
tools in language understanding and analysis of word 
behaviour~\cite{mikolov2013linguistic,bolukbasi2016man,garg2018word}. 

\begin{figure}[ht]
    \centering
    \includegraphics[width=0.95\linewidth]{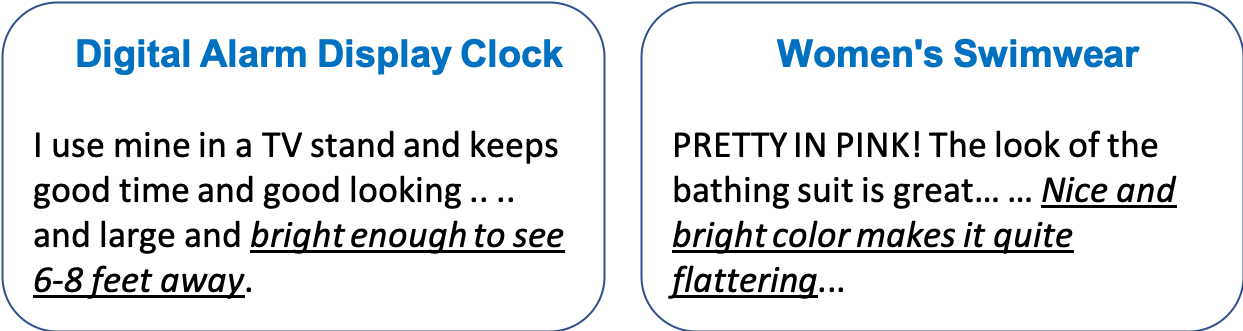}
    \caption{Sample product reviews from different domains, using the word \textit{bright} but in different contexts.}
    \label{intro-bright}
    \vspace{-0.5cm}
\end{figure}

In a typical embedding space, each word is represented using a 
single embedding vector. However, applications often involve corpora consisting of 
documents from diverse domains, where usage of a word varies across domains. 
For example, in product reviews, the word \textit{bright} usually refers 
to screen brightness for {\tt electronics} products, but when used in 
reviews of {\tt clothes} it refers to a lighter shade of colors, as shown in sample reviews in Figure \ref{intro-bright}. 
Ignoring these nuances in word representations may affect many downstream 
applications, such as sentiment analysis where 
\textit{thin} or \textit{cheap} are positive 
in the context of {\tt 
electronics}, but may be perceived as negative for {\tt outdoors} or 
{\tt apparel} products.

It is thus insufficient to learn a single representation for a word, where its  meanings and usage patterns pertaining to different topical 
domains would be lost. In order to learn separate embeddings in 
different domains, the current models \cite{mikolov2013distributed, pennington2014glove} would need to be trained independently on 
each domain-specific dataset. This raises multiple issues: (1)~not 
all domains have a large amount of data for learning a reliable embedding 
model; (2)~the relationships between domains are not captured when we treat 
them independently, and (3)~the embedding dimensions will not be aligned making 
it hard to study the semantic shifts of a word between the domains. 
Therefore, we need a principled way to learn domain-specific word embeddings 
that leverages the domain relationships.

\begin{figure}[ht]
    \centering
    \includegraphics[width=\linewidth]{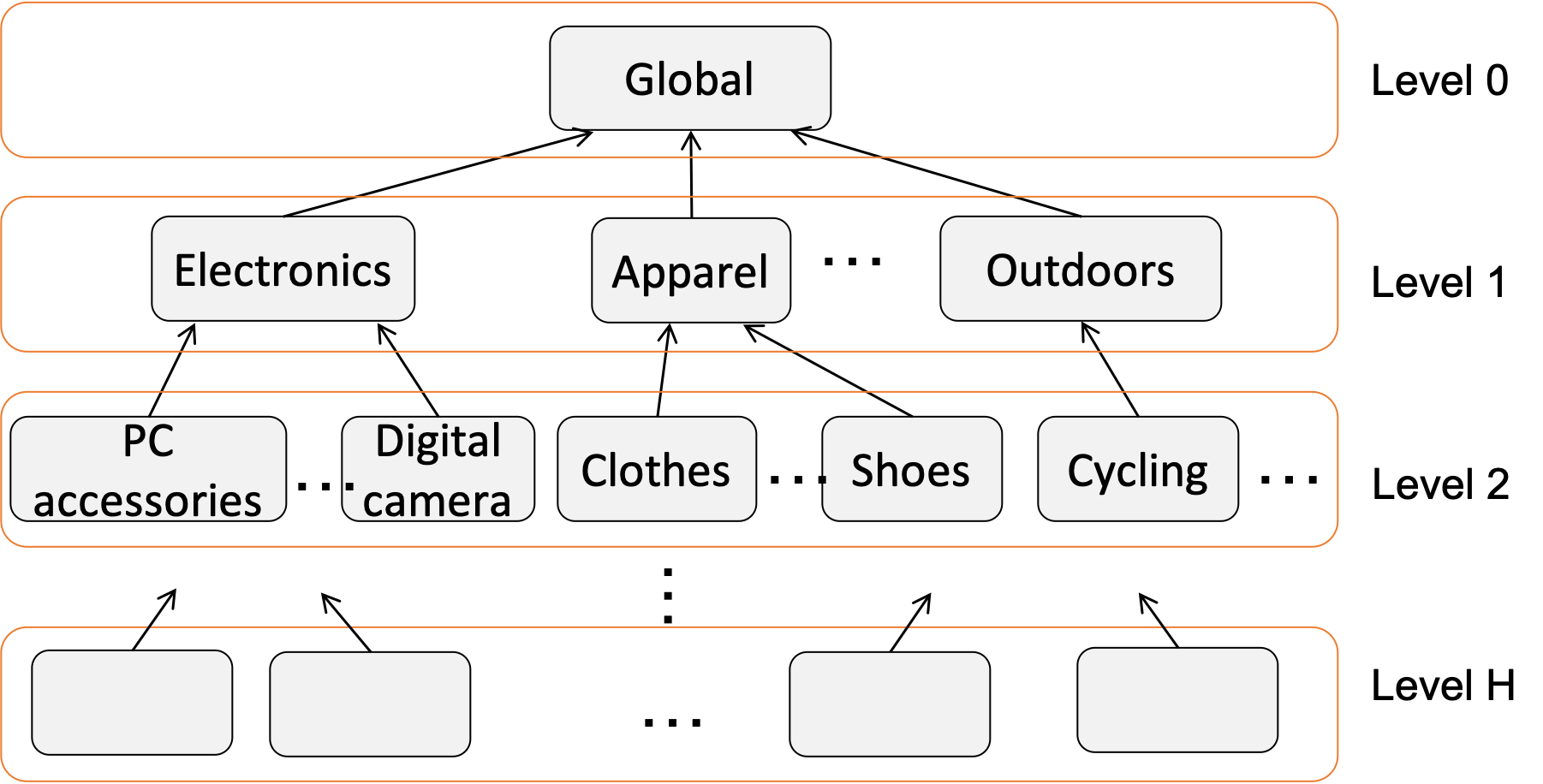}
    \caption{A sample product taxonomy where products belong to a node in the hierarchy. \textit{Global} is a dummy node representing root of the tree. }
    \label{taxonomy}
    \vspace{-0.7cm}
\end{figure}

In this paper we propose a hierarchical word embedding model that learns 
domain-specific embeddings of words from text categorized into hierarchical 
domains, such as a product taxonomy of an e-commerce site, \textcolor{black}{research field tags of scientific
articles, or topical categories} for a news portal. A sample product taxonomy is shown 
in Figure~\ref{taxonomy}. For an e-commerce site, products are categorized into 
one or more nodes in such a hierarchy. 
Products that are closer in the taxonomy (e.g.~{\tt Digital 
cameras} and {\tt PC accessories}) are more similar to each other than to 
products residing in distant nodes of the tree (e.g.~{\tt Shoes} or {\tt 
Cycling}).
We note that, while products may have 
some features specific to its particular node, they also have some features 
inherited from their parent and hence are shared across sibling nodes (e.g.,~most 
{\tt Electronic} items have a {\it battery}, or the typical product under 
{\tt Apparel} would have a {\it size}). 
This intuitively implies that most 
words in a category node show similar usage patterns to their usage in 
the parent node, while some words may deviate from the parent category-level meaning, 
and adopt specialized meanings. 
Accurate modeling of these words will not only benefit NLP systems, but 
will also facilitate discovery of such terms in a systematic way.

We build on structured Exponential Family 
Embeddings (s-EFE) \citep{rudolph2017structured}.
We assume that the embedding of a word for a node in the taxonomy, may deviate from its embedding in the parent node to a certain extent following a probability distribution.
We use the hierarchical structure of the domains to control for the amount of 
deviation allowed for a word among domains. This will help the model share
information among related product domains and allow the embedding for the same word to vary for distant domains proportional to their distance in the taxonomy.

We conduct extensive experiments on $9$ large and diverse product reviews  datasets 
and show that our model is able to fit unseen data better than competitive baselines. 
The learned semantic space can capture idiosyncrasies of different domains which are intuitive to humans, suggesting the usefulness of the representations for downstream exploratory applications. 
Evaluation on a downstream task of review rating prediction demonstrates that the hierarchical word embeddings outperform other word embedding approaches. Further analysis through crowd-sourced evaluation shows that our model can discover key domain terms, as a natural by-product of its construction without requiring additional processing.

To summarize, this paper makes the following contributions:
\begin{itemize}
 \item Proposing a hierarchical word embedding model to represent domain-specific 
word meanings, by leveraging the inherent hierarchical structure of domains. 
 \item Conducting extensive qualitative and quantitative evaluation, including 
downstream tasks demonstrating that our model outperforms competitive baselines.
\item Presenting the capability of our model in naturally learning domain-specific keywords and evaluating them using crowd-sourcing.
\end{itemize}

%% file: relatedWork_v3.tex
\section{Related Work}
\label{relatedWork}
Learning word embeddings has received 
substantial attention and has moved from co-occurrence 
based models~\citep{landauer1998introduction} 
to a wealth of prediction based methods, 
leveraging deep learning~\citep{mikolov2013efficient,pennington2014glove,bojanowski2017enriching} and Bayesian modeling~\citep{rudolph2016exponential}. 
All these methods learn a single representation of a word, which is insufficient for capturing usage variation across domains.
Recently, ~\newcite{rudolph2017structured} proposed s-EFEs to exploit grouping of
data points into a (flat) set of semantically coherent subgroups (e.g., 
grouping ArXiv papers by research field). Inspired from their approach, we develop a model to capture multi-level {\it hierarchical} structure in the data, and present product hierarchies on e-commerce 
platforms as a sizable real-world use case. 

In the spirit of capturing multiple meanings of a word,
our work is related to context-dependent word representations \cite{mccann2017learned,peters2018deep,devlin2018bert} and multi-sense word
embeddings~\citep{neelakantan2014efficient,li2015multisense,nguyen2017mixture, 
athiwaratkun2018probabilistic}. 
Contextualized word vectors \cite{mccann2017learned} and their deep extensions \cite{peters2018deep} aim to learn a dynamic representation of a word as internal states of a recurrent neural network, that encodes a word depending on its context. They have been proven effective for downstream tasks when used in addition to pre-trained single vector word embeddings. This line of research is complementary to ours and combining these would be an interesting direction to explore in future.
Multi-sense embeddings 
dynamically allocate further vectors to represent new senses of a word \cite{li2015multisense}. 
In contrast, our model is not designed to 
explicitly encode word senses, i.e. generally accepted distinct meanings of a word. We capture idiosyncrasies in word usages which are specific to the hierarchical domain structure exhibited by the data.

Our model can capture domain-specific semantics of product categories, and relates to unsupervised~\cite{titov2008modeling,poddar2017author,luo2018extra} or domain knowledge-based~\cite{mukherjee2012aspect,chen2013exploiting} models of aspect 
discovery. Aspects are 
domain-specific concepts along which products are 
characterized and evaluated (e.g., {\it battery life} for electronics, or 
{\it suspense} for crime novels). We focus on learning a 
characterization of each word specific to every domain, 
instead of learning abstract aspects. However, we show in an experiment with human 
subjects that the learnt embeddings can be utilized to discover important domains terms without further processing and suggest interesting applications for future research.


%% file: methods_v6.tex
\section{Method}
\label{method}
In this section  we first describe the  Exponential Family 
Embeddings and its extension for grouped data. Thereafter, we describe the construction of our 
hierarchical word embeddings model.

\subsection{Background}

{\bf Exponential Family Embeddings} (EFE; \newcite{rudolph2016exponential}) is a generic model that 
encodes the conditional probability of observing a data point (e.g. word in a 
sentence, item in a shopping basket) given other data points in its context 
(surrounding words, other items in the shopping basket, respectively). 

In the context of texts, a sentence consisting of $n$ words is represented as a binary matrix $X \in 
\mathbb{R}^{n \times V}$ where V is the size of vocabulary and $x_{iv}$ 
indicates whether the $v^{th}$ word appears at the $i^{th}$ position in the 
sentence or not.  EFE models the conditional distribution of an observation
$x_{iv}$, given its context $c_i = \{ j \neq i | i-w \le j \le i+w \}$ 
where $w$ is the size of the context window.
\begin{equation}
x_{iv} | \mathbf{x}_{c_{iv}} \sim ExpFam(\eta_v(\mathbf{x}_{c_{iv}}) , t(x_{iv}))
\end{equation}


\noindent where $\mathbf{x}_{c_{iv}}$ are the context vectors of $x_{iv}$; 
$\eta_v(\mathbf{x}_{c_{iv}})$ is the natural parameter, and $t(x_{iv})$ is the sufficient statistics for exponential families.

The 
natural parameter
is parameterized 
with two vectors: (1) embedding vector of a word 
$\boldsymbol{\rho}_{iv} \in \mathbb{R}^{K}$, and (2) context vectors of surrounding 
words $\boldsymbol{\alpha}_{jv^{'}} \in \mathbb{R}^{K}$; where $j \in c_i$ and $K$ is the embedding dimension.
The definition of the natural parameter is a modeling choice and for EFE it is 
defined as a function of a linear combination of the above two vectors for 
focusing on linear embeddings.  
\begin{equation}
\eta_v(\mathbf{x}_{c_{iv}}) = f(\boldsymbol{\rho}_{iv}^T \sum_{j \in c_i} 
\boldsymbol{\alpha}_{jv^{'}}x_{jv^{'}})
\end{equation}
\noindent where $f(.)$ is the identity link function.

EFEs define a \textit{parameter sharing structure} across observations. By assuming that for each word $v$ in the 
vocabulary, $\boldsymbol\rho_{iv} = \boldsymbol\rho_v$ and 
$\boldsymbol\alpha_{iv} = \boldsymbol\alpha_v$, the vectors are shared 
across positions of its appearance in a sentence.

EFEs are a general family of probabilistic graphical models that are also applicable to data 
other than language, and provide the flexibility of choosing the appropriate 
probability distribution depending on the type of data at hand. For word embeddings, we model binary word count data by observing whether or not 
a word appears in a given context i.e. $x_{iv}$ is $0$ or $1$. 
The Bernoulli distribution is a suitable choice for this type of data. 

Since the matrix $X$ is very sparse, in order to diminish the effect of negative observations and to keep the training time low, a negative sampling approach is used similar to Word2Vec \cite{mikolov2013distributed}.
The model is trained by maximizing the following objective function,

\begin{equation}
\begin{aligned}
    \mathcal{L} = &log P (\boldsymbol\alpha) + log P(\boldsymbol\rho) + \\ &\sum_{v,i} log P (x_{iv} | 
\mathbf{x}_{c_{iv}}; \boldsymbol\alpha , \boldsymbol\rho ).
\end{aligned}
\end{equation}

\noindent {\bf Structured Exponential Family Embeddings}~(s-EFE; \newcite{rudolph2017structured}) extends EFE to model data organized into a set of groups and learn group-specific embeddings for each word. 
Similar to EFE, s-EFE assumes that a word $v$ has a context 
vector $\boldsymbol\alpha_v$ and an embedding vector $\boldsymbol\rho_v$. 
\textcolor{black}{The context vector of a word ($\boldsymbol\alpha_v$) is shared 
across groups, but embedding vectors ($\boldsymbol\rho^g_v$) are 
learned specific to each group $g$. 
By sharing the context vectors ($\boldsymbol{\alpha}$), s-EFE ensures that the embedding dimensions are aligned across groups and are directly comparable. 
The embedding vectors ($\boldsymbol{\rho}$) are tied together to share statistical strength through a common prior: each $\boldsymbol\rho^g_v$ is drawn from a normal 
distribution centered around the word's global embedding vector $\boldsymbol\rho^0_v$; 
i.e. $ \boldsymbol\rho^g_v \sim \mathcal{N}(\boldsymbol\rho^{0}_v,\sigma^2) $, where 
$\sigma$ is a parameter to control how much the group vectors can deviate from the global representation of the word. }


\subsection{Hierarchical Embeddings}
The s-EFE 
model~\cite{rudolph2017structured} can learn word embeddings only over a flat list of groups, where the degree of similarity between groups are not captured. However, in a complex real-world scenario, data is often organized into a hierarchy of domains such as a product taxonomy (as shown in
Figure~\ref{taxonomy}), where each product and thereby all its reviews are 
categorized into one of the nodes. The hierarchy represents an inherent similarity measure among the domain nodes, where nodes that are closer to each other in the hierarchy, represent more similar domains, than nodes that are further away. This gets more pronounced as we continue to go deeper in the hierarchy, for e.g. \texttt{Electronics/Gaming/XBox} has a lot more in common with \texttt{Electronics/Gaming/Playstation} than with \texttt{Apparel/Shoes/Sneakers}. We believe that these similarities among the domains are also reflected in word usages in those domains i.e. a word is more likely to be used for conveying similar meaning in two closely related domains than in two distant domains in the hierarchy.

We build upon the s-EFE
model~\cite{rudolph2017structured} and extend it to learn word embeddings for 
data organized in hierarchical domains. We assume this hierarchy or taxonomy to be 
given, which we believe is a natural assumption in many real world scenarios. For a node $d$ (e.g. \texttt{Shoes}) in the hierarchy, we denote its parent node as $parent(d)$ 
(i.e. \texttt{Apparel}). For each word, we learn a specific representation 
at each node, which captures its properties in that particular 
domain.


In order to incorporate the domain hierarchy in the learned word embeddings and share information
between domains along the tree structure, we tie together the embedding vectors of sibling nodes:
We assume that, for each node $d$, the embedding vector 
$\boldsymbol\rho_v^d$ for word $v$, is drawn from a normal distribution 
centered at the embedding vector of the word in the \textit{parent} node of $d$ i.e.,
\begin{equation}
    \boldsymbol\rho^d_v \sim 
\mathcal{N}(\boldsymbol\rho^{parent(d)}_v,\sigma^2).
\end{equation}

\noindent where $\boldsymbol\rho^{parent(d)}_v$ is the embedding vector for word 
$v$ in the parent node of $d$, and $\sigma$ is the standard deviation for the 
normal distribution. The value of $\sigma$ 
controls how much an 
embedding at a node can differ from its parent node. If this value is 
low, the embeddings have to remain very close to their parent representations 
and as a result significant variation can not be observed across domains.

For a word $v$, $\boldsymbol\rho^0_v$ denotes its embedding at level $0$ (i.e.~the root) of the 
hierarchy. This can be considered as a global embedding of the word that 
encodes its properties across all domains. 
For nodes at level $1$, the embeddings for the words are learned conditioned on 
this global embedding, and similarly for any level $h$ the embeddings are learned conditioned on their parent node's embedding at level $h-1$.

This structure implies that the amount by which the embedding of a word at a node may vary from its global embedding, is proportional to the depth of the node in the hierarchy, i.e. how specific or fine-grained the domain is. This captures the intuition that fine-grained domains might have specific word usages that the embedding vectors need to accommodate. Additionally, by allowing the embeddings at a node to deviate only by a limited measure (controlled by $\sigma$) from the embedding at its parent node, our model ensures that the inherent similarities between  parent-child nodes and the sibling nodes are preserved. On the other hand, for distant nodes, the representation of the same word will be able to vary considerably -  proportional to the distance of the two domains in the taxonomy.

\begin{table*}[t]
\centering
\resizebox{\textwidth}{!}{%
\begin{tabular}{l|c|c|c|c|c|c|c|c|c|c}
 & Home & Electronics & Apparel & Camera & Kitchen & Outdoors & Sports & Furniture & Books & Total \\ \hline
\#reviews & 6.2M & 3.1M & 5.8M & 1.8M & 4.8M & 2.3M & 4.8M & 791K & 10.2M & 40M \\ 
\#sentences & 26.7M & 17.9M & 21.6M & 11.37M & 24M & 11.5M & 21.8M & 4.1M & 64.1M & 203M \\ 
\#words & 277M & 214M & 201M & 140M & 268M & 131M & 233M & 44.5M & 892M & 2.4B \\ 
Avg. sentence length & 10.37 & 11.95 & 9.31 & 12.31 & 11.16 & 11.39 & 10.68 & 10.85 & 13.93 & 11.81 \\ \hline
\end{tabular}
}
\vspace{-0.2cm}
\caption{Statistics of the review datasets.}
\label{dat_stats}
\vspace{-0.4cm}
\end{table*}

For words that do not change in meaning or are not very 
frequent in a domain, its domain-specific embedding would not need to deviate much from the global 
embedding. This should be true for most words, as not all words exhibit domain 
dependent behavior. However, for words that have a different meaning in a subcategory, 
their embeddings will deviate more strongly from the embedding at parent 
nodes in order to reflect the domain meaning. We demonstrate and analyze this effect by investigating such words in Section~\ref{disc:awd}.

We maximize the following objective to train the hierarchical model,
\begin{equation}
 \begin{aligned}
    \mathcal{L} = &log P (\boldsymbol\alpha) + log P(\boldsymbol\rho^0) + \\
& \sum_{h=1}^H \sum_{d \in h} \Big( \sum_v log 
P(\boldsymbol\rho^d_v|\boldsymbol\rho^{\text{parent(d)}}_v)  + \\
& \sum_{v,i} log P (x_{iv} | \mathbf{x}_{c_{iv}}, \boldsymbol\alpha ,
   \boldsymbol\rho^d ) \Big).
 \end{aligned}
\end{equation}

\noindent where $H$ is the maximum depth of the tree, and $d$ is a domain at level $h$. The objective function sums the log conditional probabilities of each data
point ($log P(x_{iv} | \mathbf{x}_{c_{iv}}, \boldsymbol\alpha ,
   \boldsymbol\rho^d )$), log conditional probabilities of embeddings at each domain node ($log P(\boldsymbol\rho^d_v|\boldsymbol\rho^{\text{parent(d)}}_v)$), regularizers for the global embeddings ($log P(\boldsymbol{\rho^0})$) and context vectors ($log P(\boldsymbol{\alpha})$).

%% file: experiments_v6.tex
\section{Experiments}


We consider a large collection of real-world publicly available customer review datasets from Amazon\footnote{\url{https://s3.amazonaws.com/amazon-reviews-pds/readme.html}}. We select 
datasets from $9$ popular and diverse categories, namely, {\tt Apparel, Books, Camera, Kitchen,
Electronics, Furniture, Outdoors, Home} and {\tt Sports}. For each of these categories the number of reviews varies between $\sim 800K - 10.2M$, and are written by customers over a period of two decades.  
Table \ref{dat_stats} shows the overall statistics.


We consider these $9$ categories as level $1$ nodes in a product hierarchy. 
We further map each review to a finer category corresponding to a child node in the taxonomy  (e.g., reviews for the level $1$ node {\tt books} 
would be split into review sets for {\tt romance books}, {\tt cook books}, 
etc.). We refer to these finer categories as level $2$ nodes.
In our experiments we consider this 3-level taxonomy (a global root, and two levels of product nodes, see Figure~\ref{taxonomy} for example). In principle our model can scale to deeper hierarchies.


We apply standard pre-processing techniques on review texts.
We consider the top $50K$ most frequent words as the vocabulary for learning embeddings and remove all words that are not part of the vocabulary. Following \newcite{mikolov2013distributed}, we down-sample the most frequent words by removing tokens with probability $1 - 
\sqrt{10^{-5}/f_v}$, where $f_v$ is the frequency of word $v$.

\subsection{Parameter Settings}
The depth and width of the taxonomy tree may grow as more diverse and new products are 
added to the catalogue. 
Therefore, deciding the maximum depth of a branch at which to learn embeddings becomes an implementation and scalability choice. In our experiments, we recursively merge smaller leaf nodes to their parent 
until each node has a minimum data size of $30K$ reviews. In total we had $107$ groups across level 1 and level 2 to train our hierarchical model. For products tagged with multiple categories, we consider its reviews to be associated with all those domains.

For all competing methods, we use an embedding dimension of $100$. While training the exponential family based models, we use a context window of $4$, and positive to negative ratio of $1:10$ through random negative sampling. We experiment with multiple values of the standard deviation ($\sigma$) and find that for values in the range of $\{10, 100\}$, domain variations are  captured well. We report results with $\sigma$ set to $10$.

\subsection{Quantitative Evaluation}
\label{ssec:quant-e}
We first evaluate the effectiveness of the proposed
hierarchical word embedding model compared to recent alternatives for modeling 
unseen data.
We use log-likelihood on held out data for measuring 
intrinsically how well a model is able to generalize. The higher the log-
likelihood value on unseen data, the higher is the generalization power of the 
model.
We compare with the following:

\smallskip
\noindent \textbf{Global Embedding (EFE)} \cite{rudolph2016exponential} 
This model is fit on the whole dataset and learns a single 
global embedding for a word which does not take the domains into account. 

\smallskip
\noindent \textbf{Grouped Embedding (s-EFE)} \cite{rudolph2017structured}  For the grouped model, we merge all reviews under a sub-tree to the level 1 node (e.g. \texttt{Electronics/PC 
Accessories}, \texttt{Electronics/Digital Cameras} are all merged to \texttt{Electronics}).

\begin{table}[t]
\centering
\resizebox{0.95\linewidth}{!}{
\begin{tabular}{|l|c|c|}
\hline
Model & Val & Test  \\ \hline
EFE \cite{rudolph2016exponential} & -2.015  & -2.016  \\ \hline
s-EFE \cite{rudolph2017structured} & -1.635  & -1.656  \\ \hline \hline 
Hierarchical Embedding & \textbf{-1.416} & \textbf{-1.425} \\ \hline
\end{tabular}
}
\vspace{-0.1cm}
\caption{Comparison of log-likelihood(LL) results. Higher is better. All comparisons are statistically significant (paired t-test with p $<$ 0.0001)}
\vspace{-0.4cm}
\label{logLL}
\end{table}

We divide the dataset in train ($80\%$),  validation ($10\%$), and test ($10\%$) sets. 
We learn the models on the train set and tune hyper-parameters on the validation set.
For five such random partitions we note accuracy on the validation and test set. 
As the datasets are sufficiently large,  we do not observe much fluctuation across different partitions (standard deviation  $ \le 0.0001$).
Therefore, we report
the average accuracy among the runs across all categories in Table \ref{logLL}.

As we can see from the results, the Hierarchical Embedding approach is able to 
best fit the held-out data by a significant margin. The Grouped Embedding model 
is also able to outperform the Global Embeddings. This shows that word usages do 
vary across domains, rendering a global model insufficient to generalize to 
unseen data from different domains. By outperforming the Grouped model, the 
Hierarchical model demonstrates its efficiency in being able to incorporate the 
domain taxonomy well and the effectiveness of our approach for modeling such 
hierarchical data.

\subsection{Rating Prediction}
As embeddings are popularly used as word representations for NLP applications, we evaluate the quality of hierarchical word vectors learned by our proposed approach in the context of downstream tasks. We consider the traditional task of rating prediction from 
review content. 
We first pre-train all competing embedding methods on the review dataset 
and use the trained word vectors in the following neural model for text representation.

\begin{figure}[htbp]
    \centering
    \includegraphics[width=0.95\linewidth]{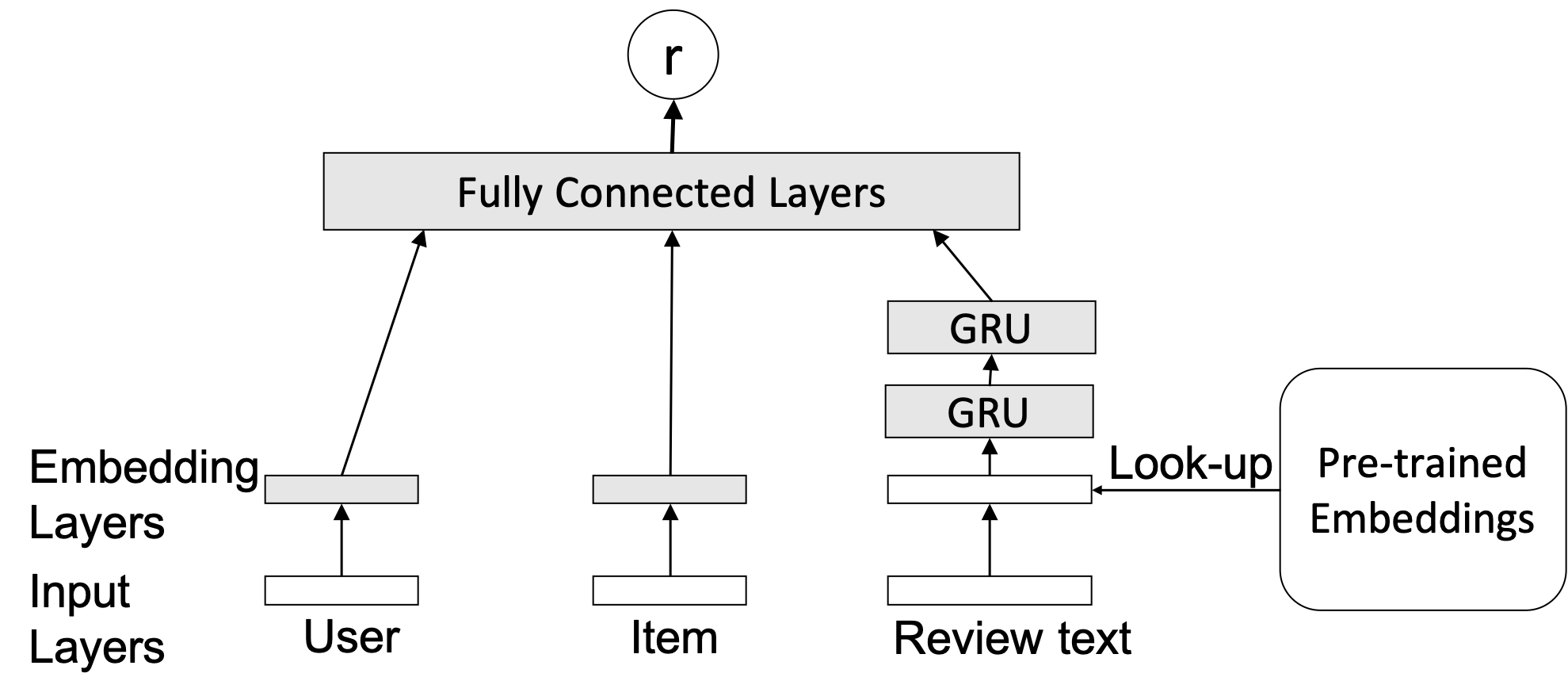}
    \caption{Architecture of the neural net model for  rating prediction. Layers 
whose weights are learned as part of the network are shaded in gray.}
    \label{ratingPred}
    \vspace{-0.2cm}
\end{figure}

\noindent\textbf{Rating Prediction Model:}  For a review, we consider the associated user id, item id and the review text in order to 
predict the star rating given by the user for the item. The model (shown in Figure \ref{ratingPred}) uses these three signals to predict a numeric rating. 

We embed the user and item id similar to 
Neural Collaborative Filtering (NeuMF) \cite{he2017neural}. For the review text, we first encode each word using different competitive pre-trained word embeddings for comparison. Thereafter, two  GRU layers ~\cite{chung2014empirical} are used to encode the review text from the sequence of word 
embeddings. 
The output from the final timestep of the GRU is considered as the encoded 
representation of review text. We concatenate the three vectors and feed them through a fully connected layer with {ReLU} activation to 
predict the numeric rating. The network is trained using {mean squared 
error} loss through backpropagation. 

\begin{table}[t]
\centering
\resizebox{\linewidth}{!}{
\begin{tabular}{|l|c|c|}
\hline
Method & MAE & RMSE \\ \hline
NeuMF (No Text) \cite{he2017neural} & 0.746 & 1.09\\ \hline
EFE \cite{rudolph2016exponential} & 0.573 & 0.874\\ \hline
s-EFE \cite{rudolph2017structured} & 0.565 & 0.878 \\ \hline
GloVe \cite{pennington2014glove} & 0.566 & 0.863\\ \hline
Word2Vec - Skip-gram \cite{mikolov2013efficient} & 0.554 & 0.854 \\ \hline
Word2Vec - CBOW \cite{mikolov2013efficient} & 0.556 & 0.847 \\  \hline \hline
Hierarchical Embedding & \textbf{0.526} & \textbf{0.843} \\
\hline
\end{tabular}%
}
\vspace{-0.1cm}
\caption{Comparison of embedding approaches for predicting Star Rating of
reviews. Lower is better.}
\label{rating-pred-reco}
\vspace{-0.5cm}
\end{table}

\begin{table*}[t]
\centering
\small
\begin{tabular}{|l|l|l|l|l|l|l|}
\hline
\texttt{Apparel} & \texttt{Camera} & \begin{tabular}[c]{@{}l@{}} \texttt{Camera/}\\ \texttt{lenses} \end{tabular} & \begin{tabular}[c]{@{}l@{}} \texttt{Camera/} \\ \texttt{flashes} \end{tabular} & \begin{tabular}[c]{@{}l@{}} \texttt{Books/} \\\texttt{cookbooks} \end{tabular} & 
\begin{tabular}[c]{@{}l@{}} \texttt{Books/} \\ \texttt{travel}\end{tabular} & \begin{tabular}[c]{@{}l@{}} \texttt{Kitchen/} \\ \texttt{bakeware} \end{tabular}\\
\hline
jegging & digital & blurriness & pandigital & pollan & pilgrimage & springfoam 
\\
bikini & zoom & focusing & speedlights & foodies & shrines & corningware \\
headwear & zooming & hsm & jittery & vegetarian & crossings & pizzeria \\
backless & nikor & focal & nikor & lentil & kauai & basting \\
danskin & pixel & aperture & jpegs & healthful & regional & tinfoil \\
sunbathing & cybershot & nightshot & closeups & crocker & geology & molds \\
necks & washout & af & webcams & tacos & hiker & microwaveable \\
coolmax & ultrahd & fringing & studio & turnip & ecosystem & muffins \\
jockey & fringing & monopod & lamps & pastes & guides & browned \\ \hline
\end{tabular}
\vspace{-0.2cm}
\caption{Top words with most deviations in a domain reflect salient domain terms. `` / '' is the level delimiter.}
\label{domain-words}
\vspace{-0.5cm}
\end{table*}

\noindent \textbf{Baselines:} We evaluate our hierarchical embedding model against Global 
Embeddings (EFE) and Grouped Embeddings (s-EFE). We also compare with 
Word2Vec~\cite{mikolov2013efficient} (both Continuous Bag of Words (CBOW) and
Skip-gram, trained on our reviews corpus) and GloVe~\cite{pennington2014glove} (pre-trained on Wikipedia+Gigaword) which are among the most commonly used 
 word embedding approaches. For Word2Vec, we compare with both the Continuous Bag of Words (CBOW) and Skip-gram algorithm using the implementations from Python's Gensim library \cite{rehurek_lrec}.

We split the dataset in 80-10-10 proportions for train-development-test and report 
mean absolute error (MAE) and root mean squared error (RMSE) over test set 
after averaging five runs with random partitions.
Table \ref{rating-pred-reco} shows the results of using different embeddings 
for the rating prediction task. 
We can see that all 
methods that include text perform better than NeuMF which only uses the user id 
and item id. This shows the benefit of utilizing the textual information.
We also observe that both of the Word2Vec algorithms outperform GloVe embeddings, demonstrating the domain dependence of word meanings that are not captured in GloVe embeddings pre-trained on Wikipedia and Gigaword.
Finally, the proposed Hierarchical Embeddings provide the best text representation and achieve the lowest error rate. By learning 
domain specific word embeddings, the hierarchical model can represent 
appropriate usage of a word, leading to better downstream predictions.

\subsection{Domain Term Discovery}
\label{disc:awd}
Discovering key domain specific terms is an important building block for 
many downstream applications of review content understanding. 
For example, while \textit{aperture}, \textit{focus}, 
\textit{sharpness} are aspects of camera lenses,  words such as \textit{size}, 
\textit{full-sleeve},  \textit{turtleneck} and so on are important 
\textcolor{black}{to characterize a fashion product.} 
\textcolor{black}{Mining} such domain terms can help in fine grained sentiment analysis 
and can facilitate the study of customer opinions at a detailed, actionable level 
or help in extracting accurate feature specifications from product descriptions,
improve indexing and product discovery and many such downstream tasks. 
However, with constant influx of new products from diverse domains, it is infeasible to manually curate such terms with great detail and coverage.

Customer reviews can be a rich source of information 
for discovering \textcolor{black}{domain terminology}. 
They discuss the properties of products or 
businesses that customers truly care about.
We explore the potential of our embedding approach 
for discovering such words in a data-driven fashion.

In the Hierarchical Embedding model, the embedding of a word in a domain may differ from the one in the parent domain following a distribution. For domain-specific words we assume that their embeddings differ \textit{more} 
from the embeddings at the parent or the global embeddings than for 
domain-neutral words, in order to accommodate the domain usage. 
We study our model's ability to discover domain terms by inspecting the words whose embeddings deviated the most from their parent domain.

Table \ref{domain-words} shows the ranked list of top $10$ such words for a few domains. We can 
observe that most of the words with highest embedding deviations in a domain are 
indeed domain-specific. This shows that the hierarchical embedding approach is able to capture usage variation of words and can help discover  \textcolor{black}{salient domain-specific words without requiring any further processing steps. Additionally, synonyms of these words can 
be explored among their neighboring words in the embedding space to augment the domain term list.}

\begin{table}[tbp]
    \centering
    \resizebox{0.85\linewidth}{!}{
    \begin{tabular}{|l|c|c|}
    \hline
    Domain Levels & \#Pairs     & Accuracy \\ \hline
    Level 1  & 81 & 97.5\% \\
    Level 2 (different parents) & 491 & 95.71\% \\
    Level 2 (same parent) & 491 & 82.45\% \\ \hline
    \end{tabular}
    }
    \vspace{-0.1cm}
    \caption{Accuracy of detecting the domain given the top most deviating 
words for different domain levels.}
    \label{aspect_mturk}
    \vspace{-0.5cm}
\end{table}

To quantitatively evaluate how well these words can describe a domain, we set up 
an annotation task on Amazon Mechanical Turk\footnote{\url{https://www.mturk.com/}}. 
Given a set of $5$ words, we ask MTurk workers to choose among a pair of domains, which domain the words belong to. 
We perform the experiment at multiple levels of 
granularity and 
select the pair of domains from- (1) two level $1$ nodes (e.g. \texttt{Apparel} vs. \texttt{Electronics}), (2) two level $2$ nodes from different parents (e.g. \texttt{Electronics/Homedecor} 
vs. \texttt{Home/Living room furniture}), and (3) two level $2$ nodes from the same parent (e.g. \texttt{Books/Travel} vs. \texttt{Books/Science}).

\begin{table*}[t]
\centering
\resizebox{\textwidth}{!}{
\begin{tabular}{|l|l|l|l|l|l|}
\hline
\multicolumn{2}{|l|}{\begin{tabular}[c]{@{}l@{}}\textbf{\texttt{Kitchen}}\\ razor, dull, cutting, sharpen, sharpness\end{tabular}} & \multicolumn{2}{l|}{\begin{tabular}[c]{@{}l@{}}\textbf{\texttt{Outdoor}} \\ edges, edge, sharpened, cuts, flakes\end{tabular}} & \multicolumn{2}{l|}{\begin{tabular}[c]{@{}l@{}}\textbf{\texttt{Camera}}\\ stunningly, combined, compensated, distinctly, combining\end{tabular}} \\ \hline
\begin{tabular}[c]{@{}l@{}}\textbf{\texttt{Cookware}}\\ razor, dull, cutting,\\ knife, knives\end{tabular} & \begin{tabular}[c]{@{}l@{}}\textbf{\texttt{Cutlery and}}\\ \textbf{\texttt{knife accessories}} \\ razor, effortless,\\ strokes, dull, stab\end{tabular} & \begin{tabular}[c]{@{}l@{}}\textbf{\texttt{Cycling}}\\ leaves, glass, cuts,\\ hits, stones\end{tabular} & \begin{tabular}[c]{@{}l@{}}\textbf{\texttt{Accessories}}\\  action, jacket,\\ dust, hat, wind\end{tabular} & \begin{tabular}[c]{@{}l@{}}\textbf{\texttt{Flashes}}\\  silky, crystal, clear, \\ nuance, perfection\end{tabular} & \begin{tabular}[c]{@{}l@{}}\textbf{\texttt{Lenses}}\\  yielding, useable, enlarged,  pixelation,\\ enlarge\end{tabular} \\ \hline
\end{tabular}
}
\vspace{-0.2cm}
\caption{Top neighboring words for \textit{sharp} across different domains}
\label{demo-hier-sharp}
\vspace{-0.2cm}
\end{table*}

\begin{table*}[t]
\centering
\resizebox{\textwidth}{!}{
\begin{tabular}{|l|l|l|l|l|l|}
\hline
\multicolumn{2}{|l|}{\begin{tabular}[c]{@{}l@{}}\textbf{\texttt{Electronics}}\\ light, lights, dark, green, lit\end{tabular}} & \multicolumn{2}{l|}{\begin{tabular}[c]{@{}l@{}}\textbf{\texttt{Outdoor}} \\ practice, effort, dark, fill, darker\end{tabular}} & \multicolumn{2}{l|}{\begin{tabular}[c]{@{}l@{}}\textbf{\texttt{Camera}} \\ sun, shade, darkness, dark, lit\end{tabular}} \\ \hline
\begin{tabular}[c]{@{}l@{}}\textbf{\texttt{Computer}}\\ \textbf{\texttt{accessories}} \\ dark, light, glow,\\ green, eyes\end{tabular} & \begin{tabular}[c]{@{}l@{}}\textbf{\texttt{Homedecor}} \\ brighter, dimly, brightest, \\ levels, unobtrusive\end{tabular} & \begin{tabular}[c]{@{}l@{}}\textbf{\texttt{Cycling}}\\ steady, brighter, visible, \\ modes, flashlight\end{tabular} & \begin{tabular}[c]{@{}l@{}}\textbf{\texttt{Outdoor}} \\ \textbf{\texttt{clothing}} \\ white, purple, \\dark, actual, dress\end{tabular} & \begin{tabular}[c]{@{}l@{}}\textbf{\texttt{Flashes}} \\ balanced, spotlight,\\ reflect,  artificial, glow\end{tabular} & \begin{tabular}[c]{@{}l@{}}\textbf{\texttt{Lenses}} \\ darkness, lights,\\ direct,  brighter, sun\end{tabular} \\ 
\hline
\end{tabular}
}
\vspace{-0.2cm}
\caption{Top neighboring words for \textit{bright} across different domains}
\label{demo-hier-bright}
\vspace{-0.5cm}
\end{table*}

We evaluate on $1063$ such pairs and ask $10$ workers to evaluate each pair. 
Table \ref{aspect_mturk} shows that overall, the workers could 
identify the correct domain with high accuracy by looking only at the top $5$ words. 
As anticipated, we observe that it is easier to distinguish between more 
heterogeneous nodes (level 1 nodes or level 2 nodes with different parents) 
compared to closely related ones(level 2 nodes with same parent). 
Analyzing the mistakes made by workers, we realize that some of the finer 
domains, are indeed harder to distinguish from 
word usage as they are similar in nature (e.g. \texttt{Kitchen/Bakeware} vs. 
\texttt{Kitchen/utilities}) or are overlapping (e.g. 
\texttt{Electronics/Computer accessories} vs. \texttt{Electronics/Headphones}). 

\subsection{Qualitative Evaluation}
Finally, we qualitatively analyze whether the learned semantic space by our model is interpretable to humans for distinguishing word usages across domains. 
In the embedding space, the nearest neighbors of a  word represent its most semantically similar words. Table \ref{demo-hier-sharp} and Table \ref{demo-hier-bright} show the $5$ most similar words (by euclidean distance between word vectors) for sample words \textit{sharp} and \textit{bright} respectively, across a few domains.

Table \ref{demo-hier-sharp} shows that similar words for \textit{sharp} 
vary widely among the domains. For products in the {\tt Kitchen} domain, 
words related to \textit{sharp} are \textit{knives, sharpen, dull, razor} etc., 
whereas, for products related to {\tt Outdoor} sports and gear, \textit{sharp} is 
often used in the context of sharp wind, sharp edges or sharp pain. In the context of 
{\tt Camera}, \textit{sharp} mostly refers to image quality. We can observe that the 
hierarchical model is able to capture nuances even at a finer domain level of 
{\tt Camera/Flashes} vs. {\tt Camera/Lenses}. When used in reviews of 
{\tt Camera Flashes}, people describe the quality of the flash light in 
producing a clear, well illuminated image hence \textit{ clear, silky, 
perfection} are its most similar words. Whereas in reviews of {\tt Camera Lens}, people 
use \textit{sharp} in the context of image resolution,  describing the effect of enlargement and 
pixelation of the picture, hence related terms are captured.

Similarly, usage variation of the word \textit{bright} is captured (Table 
\ref{demo-hier-bright}). In {\tt 
Electronics} products, 
\textit{bright} is an attribute for screen display or LED lights on different 
devices. For {\tt Outdoor} sports such as \textit{cycling}, it is used in the 
context of visibility but for {\tt Outdoor Clothing}, \textit{bright} is a 
descriptor for the color of a clothing item. In reviews of {\tt Camera 
Flashes}, people refer to the artificial brightness created by the camera 
flashlight. In contrast, in reviews of {\tt Camera Lenses}, people usually discuss 
the capability of the lens to capture images under a bright sun or low-light conditions. Neighborhoods of \textit{bright} reflect this usage variation across domains.

%% file: conclusion_v2.tex
\section{Conclusion}
We have studied word embeddings to address varying word usages across domains.
We propose a hierarchical embedding that uses 
probabilistic modeling to learn domain-specific word representations, leveraging the inherent hierarchical structure of data, such 
as reviews in e-commerce site following a product taxonomy.
Our principled approach enables the learned embeddings to capture domain similarities and as
the embedding dimensions are aligned across domains, it facilitates interesting studies of semantic shifts of word usage. 
On large real-word product review datasets, we show that our nuanced representations, 
(1) provide a better intrinsic fit for the data, (2) lead to 
an improvement in a downstream task of rating prediction over state-of-the-art approaches, and (3) are intuitively meaningful to humans, opening up avenues for future explorations on aspect discovery.